\documentclass{article}

\usepackage{times}
\usepackage{graphicx} 

\usepackage{natbib}

\usepackage{algorithm}
\usepackage{algorithmic}

\usepackage[utf8]{inputenc} 
\usepackage[T1]{fontenc}    
\usepackage{hyperref}       
\usepackage{url}            
\usepackage{booktabs}       
\usepackage{amsfonts}       
\usepackage{nicefrac}       
\usepackage{microtype}      
\usepackage{subfig}
\usepackage{epstopdf}
\usepackage[titletoc]{appendix}
\usepackage{amsmath}
\usepackage{amsthm}
\usepackage{array}
\usepackage{color}
\usepackage{caption}
\usepackage{mathtools}
\usepackage{mathrsfs}
\usepackage{wrapfig}
\usepackage{multirow}
\usepackage{amssymb}
\usepackage{wrapfig}

\newtheorem{theorem}{Theorem}
\newtheorem{corollary}{Corollary}

\newtheorem{assumption}{Assumption}

\newtheorem{theoremappendix}{Theorem}

\DeclareMathOperator*{\argmin}{arg\,min}
\DeclareMathOperator*{\argmax}{arg\,max}

\DeclareMathOperator{\sgn}{sign}
\DeclareMathOperator{\softmax}{\mathbb{S}}
\DeclareMathOperator{\R}{\mathbb{R}}

\DeclareMathOperator{\eff}{Eff}
\DeclareMathOperator{\er}{ER}
\DeclareMathOperator{\I}{I}
\DeclareMathOperator{\GOM}{GenerateOptMeans}
\DeclareMathOperator{\clip}{clip}
\DeclareMathOperator{\stt}{s.t.}
\DeclareMathOperator{\RB}{RB}

\newcommand{\ep}{\mathbb{E}}

\newcommand{\HH}{\mathbb{H}}

\usepackage[accepted]{icml2018}
\usepackage[english]{babel}

\icmltitlerunning{Max-Mahalanobis Linear Discriminant Analysis Networks}

\begin{document}

\twocolumn[
\icmltitle{Max-Mahalanobis Linear Discriminant Analysis Networks}




\begin{icmlauthorlist}
\icmlauthor{Tianyu Pang}{to}
\icmlauthor{Chao Du}{to}
\icmlauthor{Jun Zhu}{to}
\end{icmlauthorlist}

\icmlaffiliation{to}{Dept. of Comp. Sci. \& Tech., BNRist Center, State Key Lab for Intell. Tech. \& Sys., THBI Lab, Tsinghua University, Beijing, 100084, China}

\icmlcorrespondingauthor{Jun Zhu}{dcszj@mail.tsinghua.edu.cn}


\vskip 0.3in
]



\printAffiliationsAndNotice{}  

\begin{abstract}
A deep neural network (DNN) consists of a nonlinear transformation from an input to a feature representation, followed by a common softmax linear classifier. Though many efforts have been devoted to designing a proper architecture for nonlinear transformation, little investigation has been done on the classifier part. In this paper, we show that a properly designed classifier can improve robustness to adversarial attacks and lead to better prediction results. Specifically, we define a Max-Mahalanobis distribution (MMD) and theoretically show that if the input distributes as a MMD, the linear discriminant analysis (LDA) classifier will have the best robustness to adversarial examples. We further propose a novel Max-Mahalanobis linear discriminant analysis (MM-LDA) network, which explicitly maps a complicated data distribution in the input space to a MMD in the latent feature space and then applies LDA to make predictions. Our results demonstrate that the MM-LDA networks are significantly more robust to adversarial attacks, and have better performance in class-biased classification.

\end{abstract}

\section{Introduction}
Deep neural networks (DNNs) have shown state-of-the-art performance in different tasks~\citep{Goodfellow-et-al2016}. A typical feed-forward DNN is a combination of a nonlinear transformation from the input $x$ to the latent feature vector $z$ and a linear classifier acting on $z$ to return a prediction for $x$. Many different architectures for neural networks have been proposed, e.g., VGG nets~\citep{simonyan2014very}, Resnets~\citep{He2015,He2016} and Google nets~\citep{szegedy2016rethinking} 
for powerful nonlinear transformation, while leaving the linear classifier part under-explored, which is by default defined as a softmax regression (SR) (or logistic regression (LR) for binary classification). Some work has tried to instead use linear (or kernel) SVMs as the classifier~\citep{huang2006large,ngiam2010tiled,coates2011analysis,tang2013deep}. But, such techniques either do not fine-tune the lower level features w.r.t. the SVM's objective or only result in marginal improvements. Thus, SR is still the default choice, given its simplicity and smoothness.

However, the SR (or LR) classifier is not problemless.~\citet{efron1975efficiency} shows that if the input $x$ arises from a 2-center mixture of Gaussian distribution, then LR is less efficient than linear discriminant analysis (LDA), i.e., LR needs more training samples than LDA does to obtain a certain error rate.
The relative efficiency of LR to LDA depends on the Mahalanobis distance $\Delta$ between the two Gaussian components and the log-ratio of class priors $\zeta$. Generally, a larger value of $\Delta$ or $\left|\zeta\right|$ will lead to a lower relative efficiency of LR to LDA.
Furthermore, it has been widely recognized that the DNNs with a SR classifier are vulnerable to adversarial attacks~\citep{Szegedy2013,Goodfellow2014,Nguyen2015,Moosavidezfooli2016}, where human imperceivable images can be crafted to fool a high-accuracy network.
Though many efforts have been devoted to improving the robustness, such as using adversarial training~\citep{Szegedy2013,Goodfellow2014,kurakin2016adversarial}, it still remains open on how to design a robust classifier by itself.

In this paper, we draw inspirations from Efron's analysis and design a robust classifier that is generally applicable to feedforward networks.
Specifically, we define the Max-Mahalanobis distribution (MMD) for multi-class classification, which is a special mixture of Gaussian distribution. We theoretically show that if the input samples distribute as a MMD, the LDA classifier will have the best robustness to adversarial attacks. Though distributing as a MMD is not likely to hold for complex data (e.g., images in the pixel space), it may be true if we properly transform the data. Based on this result, we propose a novel Max-Mahalanobis linear discriminant analysis (MM-LDA) network, which explicitly learns a powerful nonlinear transformation network to turn the complex inputs to match the MMD in a latent feature space, and then uses the LDA principle to make predictions on the latent features. Besides robustness, since a large value of $\left|\zeta\right|$ for the data distribution indicates a high efficiency of LDA, the MM-LDA network can also perform better on \emph{class-biased datasets},\footnote{Our setting differs from the previous work~\citep{huang2017discriminative,fallah2017novel}, where only the training set is class-biased.} i.e., datasets with different numbers of data points for different classes, which are common in practice.

Unlike the SR classifier, whose parameters are jointly learned with those of the transformation network, the optimal parameters of the MMD are estimated separately by a simple procedure, and we only need to learn the parameters of the transformation network.
The overall training objective is a cross-entropy loss. Standard training algorithms (e.g., stochastic gradient descent) are applicable with little extra computational cost.
Moreover, as the MM-LDA network differs only in the classifier part, our technique can be naturally combined with different kinds of nonlinear transformation architectures (e.g., VGG nets, Resnets or Google nets) and different kinds of training methods~\citep{liu2016large,Pang2017} for good performance.

We test the proposed network on the widely used MNIST and CIFAR-10 datasets for both robustness to adversarial attacks and classification accuracy. As for robustness, 
we consider various adversarial attacking methods, and the results demonstrate that the MM-LDA network is indeed much more robust to adversarial examples than the SR networks, even when the SR networks are enhanced by the adversarial training methods. As for classification, we test the performance of the MM-LDA network on both class-biased and class-unbiased datasets.
The results show that the MM-LDA networks can obtain higher accuracy on class-biased datasets while maintaining state-of-the-art accuracy on class-unbiased datasets.

\vspace{-.1cm}
\section{Preliminary Knowledge}
\vspace{-.1cm}
In this section, we first briefly introduce some notations. Then we provide a formal description of the adversarial setting and introduce some common attacking methods. Finally, we introduce the relative efficiency of logistic regression (LR) to linear discriminant analysis (LDA) in the binary-class cases, which inspires our novel network.

\vspace{-.1cm}
\subsection{Notations}
\vspace{-.1cm}
We refer to the DNN with a softmax output layer as an SR network, which is widely used in classification tasks~\citep{Goodfellow-et-al2016}. Let $L$ denote the number of classes ($L\geq2$), and define $[L]=\{1, \cdots, L\}$. An SR network can be generally expressed as $F(x,\theta)=\softmax(W_sz+b_s)$, where $z$ is the latent feature representation of the input $x$ and the softmax function $\softmax(z): {\R^L}\to{\R^L}$ is defined as $\softmax(z)_i={\exp(z_i)}/{\sum_{i=1}^{L}\exp(z_i)}$ for each element $i \in [L]$. Here, $\theta$ denotes the parameters of the nonlinear transformation network from $x$ to $z$. 
$W_{s}$ and $b_{s}$ are the weight matrix and bias vector of the softmax layer respectively. 

\subsection{The Adversarial Setting}\label{AS}
In the adversarial setting, adversaries apply attacking methods to craft adversarial examples based on the given normal examples. We consider the \emph{white-box attack}, which is the most challenging and difficult threat model for classifiers to defend~\citep{carlini2017adversarial}. White-box adversaries know everything, e.g., parameters, about the classifiers that they attack on. An adversarial example $x^*$ should be indistinguishable from its normal counterpart $x$ by human observers, but makes the classifier misclassify on it. Formally, the adversarial example $x^*$ crafted on $x$ should satisfy
\begin{equation}
\label{adv}
\hat{y}(x^*)\neq\hat{y}(x)\text{, }\stt \|x^*-x\|\leq \epsilon\text{,}
\end{equation}
where $\hat{y}(\cdot)$ denotes the predicted label from the classifier, and $\epsilon$ is the maximal perturbation under a norm that varies in different attacking methods. If there is an additional constraint that $\hat{y}(x^*)$ is a specific class $l$, $x^*$ is regarded as targeted. Otherwise $x^*$ is untargeted. Let $\mathcal{L}(x, y)$ denote the training loss on $(x,y)$. Some of the most common attacking methods are introduced below:

\textbf{Fast Gradient Sign Method (FGSM)}~\citep{Goodfellow2014} is an one-step attacking method that the adversarial example $x^*$ is crafted as
$x^*=x+\epsilon\cdot\sgn(\nabla_{x}\mathcal{L}(x, y))$.

\textbf{Basic Iterative Method (BIM)}~\citep{Kurakin2016} is an iterative version of FGSM. Let $x^*_0=x$, $r$ be the number of iteration steps, then BIM crafts an adversarial example as
$x^*_i=\clip_{x,\epsilon}(x^*_{i-1}+\frac{\epsilon}{r}\cdot\sgn(\nabla_{x^*_{i-1}}\mathcal{L}(x^*_{i-1}, y)))$,
where $\clip_{x,\epsilon}(\cdot)$ is the clipping function.

\textbf{Iterative Least-likely Class Method (ILCM)}~\citep{Kurakin2016} is a targeted version of BIM with the formula as
$x^*_i=\clip_{x,\epsilon}(x^*_{i-1}-\frac{\epsilon}{r}\cdot\sgn(\nabla_{x^*_{i-1}}\mathcal{L}(x^*_{i-1}, y_{ll})))$,
where $y_{ll}=\argmin_iF(x)_i$ is the label with minimal confidence.

\textbf{Jacobian-based Saliency Map Attack (JSMA)}~\citep{Papernot2015} is also a targeted attack that perturbs the feature $x_i$ by a perturbation $\epsilon$ in each iteration step that maximizes the saliency map
\[S(x,y)[i]=\begin{cases}
0\text{, if } \frac{\partial F(x)_y}{\partial x_i}<0\text{ or }\sum_{j\neq y}\frac{\partial F(x)_j}{\partial x_i}>0\text{,}\\
(\frac{\partial F(x)_y}{\partial x_i})\left| \sum_{j\neq y}\frac{\partial F(x)_j}{\partial x_i} \right|\text{, otherwise.}
\end{cases}\]
JSMA perturbs fewer pixels compared to other attacks.

\textbf{Carlini \& Wagner (C\&W)}~\citep{Carlini2016} defines $x^*(\omega)=\frac{1}{2}(\tanh({\omega})+1)$ in terms of $\omega$, and solves
$\min_\omega \|x^*(\omega)-x\|_2^2+c\cdot f(x^*(\omega))$,
where $c$ is a constant chosen by a modified binary search. Let $\softmax_{pre}(x)$ be the input vector of the softmax function in a classifier, then $f(\cdot)$ is the objective function defined as
\[f(x)=\max(\max\{\softmax_{pre}(x)_i:i\neq y \}-\softmax_{pre}(x)_i, -\kappa)\text{,}\]
where $\kappa$ controls the confidence on adversarial examples.

The attacking methods are generally gradient-based. They can be categorized into two groups. 
The first one consists of iterative-based methods, e.g., FGSM, BIM, ILCM and JSMA. These methods usually iterate less than hundreds of rounds to craft an adversarial example. Besides, they often blend the constraints into their updating operations (e.g., adding a sign function on the gradients under the $L_{\infty}$ norm constraint). The second group consists of optimization-based methods, e.g., the C\&W method. These methods require much more computation compared to the iterative-based methods, since optimization-based methods iterate thousands of rounds to craft an adversarial example. However, usually an optimization-based method has higher success rates on attacking classifiers.

\subsection{The Relative Efficiency of LR to LDA}\label{ARE}
Softmax regression (SR) is the most commonly used model as the linear classifier part in neural networks~\citep{Goodfellow-et-al2016}. In the binary-class cases, SR reduces to LR.
We denote the two classes with labels $0$ and $1$, and make the following assumption of the input $x$ with its label $y$.

\begin{assumption}\label{ass:1}
The distribution of the $p$-dimensional random vector $x$ with its class label $y$ is
\begin{equation*}
\label{eq:1}
\begin{split}
P(y=i)&=\pi_{i}\text{, }P(x|y=i)=\mathcal{N}(\mu_i,\Sigma)\text{,}\\
\end{split}
\end{equation*}
where $i\in\{0,1\}$, $\pi_{0}+\pi_{1}=1$ and each conditional Gaussian distribution has the same covariance matrix $\Sigma$.
\end{assumption}


\citet{efron1975efficiency} shows that under Assumption~\ref{ass:1}, LR is asymptotically less efficient than LDA. Specifically,  we denote the decision regions of a classifier as $R_0$ and $R_1$, the error rate of a classifier is defined as
\begin{equation*}
\begin{split}
\er&=\pi_0P(x\in R_1|x\sim\mathcal{N}(\mu_0,\Sigma))\\
&+\pi_1P(x\in R_0|x\sim\mathcal{N}(\mu_1,\Sigma))\text{.}
\end{split}
\end{equation*}
Then the relative efficiency of LR to LDA is defined as
\begin{equation*}
\eff_p(\zeta,\Delta)=\lim_{N\to\infty}\frac{\ep[\er_{\text{LDA}}-\er_{\text{Bayes}}]}{\ep[\er_{\text{LR}}-\er_{\text{Bayes}}]}\text{,}
\end{equation*}
where $\er_{\text{Bayes}}$ is the Bayes error rate, $N$ is the number of training data points and $\Delta=[(\mu_1-\mu_0)^\top \Sigma^{-1}(\mu_1-\mu_0)]^{\frac{1}{2}}$ is the Mahalanobis distance between the two conditional Gaussian components. A lower value of $\eff_p(\zeta,\Delta)$ indicates that asymptotically LDA needs less training data points than LR does to obtain a certain error rate.

In order to calculate $\eff_p(\zeta,\Delta)$, we let $A_i$ be
\begin{equation*}
A_i(\pi_0,\Delta)=\int_{-\infty}^{\infty}\frac{e^{-\Delta^2/8}x^i\varphi(x)}{\pi_0e^{-\Delta x/2}+\pi_1e^{\Delta x/2}}dx\text{,}
\end{equation*}
where $\varphi(x)=(2\pi)^{\frac{1}{2}}\exp(-x^2/2)$ is the probability density function of $\mathcal{N}(0,1)$. Then there is:

\begin{theorem}\label{the:0}
\citep{efron1975efficiency} The relative efficiency of logistic regression to linear discriminant analysis is
\begin{equation*}
\eff_p(\zeta,\Delta)=(Q_1+(p-1)Q_2)/(Q_3+(p-1)Q_4)\text{,}
\end{equation*}
where $Q_2=1+\pi_0\pi_1\Delta^2$, $Q_4=\frac{1}{A_0}$ and
\begin{equation*}
\begin{split}
Q_1&=\begin{pmatrix}
    1 & \frac{\zeta}{\Delta}
  \end{pmatrix}
  \begin{bmatrix}
    1+\frac{\Delta^2}{4} &(\pi_0-\pi_1)\frac{\Delta}{2}\\
    (\pi_0-\pi_1)\frac{\Delta}{2} & 1+2\pi_0\pi_1\Delta^2
  \end{bmatrix}
  \begin{bmatrix}
    1 \\
    \frac{\zeta}{\Delta}
  \end{bmatrix}\text{,}\\
Q_3&=\begin{pmatrix}
    1 & \frac{\zeta}{\Delta}
  \end{pmatrix}
  \frac{1}{A_0A_2-A_1^2}
  \begin{bmatrix}
    A_2 &A_1\\
    A_1 & A_0
  \end{bmatrix}
  \begin{bmatrix}
    1 \\
    \frac{\zeta}{\Delta}
  \end{bmatrix}. 
\end{split}
\end{equation*}
\end{theorem}
%
%
%
Generally, larger values of $\left|\zeta \right|$ or $\Delta$ imply lower values of $\eff_p(\zeta,\Delta)$, and thus lower relative efficiency of LR to LDA.

\section{Methodology}
We now present our method in this section. We first define the Max-Mahalanobis distribution (MMD) with theoretical analyses, and then propose the Max-Mahalanobis linear discriminant analysis (MM-LDA) network.

\subsection{Max-Mahalanobis Distribution}\label{MMD}
We consider the multi-class cases, and a natural extension of Assumption~\ref{ass:1} is as follows.
\begin{assumption}\label{ass:2}
The distribution of the $p$-dimensional random vector $x$ with its class label $y$ is
\begin{equation*}
\label{eq:2}
P(y=i)=\pi_{i}\text{, }P(x|y=i)=\mathcal{N}(\mu_i,\Sigma)\text{,}
\end{equation*}
where $i\in[L]$, $\sum_{i=1}^L\pi_{i}=1$ and each conditional Gaussian distribution has the same covariance matrix $\Sigma$.
\end{assumption}
Then, the Mahalanobis distance between any two Gaussian components $i$ and $j$ is $\Delta_{i,j}=[(\mu_i-\mu_j)^\top \Sigma^{-1}(\mu_i-\mu_j)]^{\frac{1}{2}}$.
As suggested in \citet{efron1975efficiency}, there is no loss of generality to assume that $\Sigma$ is nonsingular. Thus we can do the Cholesky decomposition as $\Sigma=QQ^\top $, where $Q$ is a lower triangular matrix with positive diagonal entries. By applying the linear transformation $\widetilde{x}=Q^{-1}(x-\overline{\mu})$, where $\overline{\mu}=\sum_{i=1}^L\mu_{i}/L$, we can reduce Assumption~\ref{ass:2} to the standard form.
\begin{assumption}\label{ass:3}
The distribution of the $p$-dimensional random vector $x$ with its class label $y$ is
\begin{equation*}
\label{eq:3}
P(y=i)=\pi_{i}\text{, }P(\widetilde{x}|y=i)=\mathcal{N}(\widetilde{\mu}_i,\I)\text{,}
\end{equation*}
where $i\in[L]$, $\sum_{i=1}^L\pi_{i}=1$ and $\sum_{i=1}^L\widetilde{\mu}_i=0$.
\end{assumption}
For the standard form, we have $\widetilde{\Delta}_{i,j}=[(\widetilde{\mu}_i-\widetilde{\mu}_j)^\top (\widetilde{\mu}_i-\widetilde{\mu}_j)]^{\frac{1}{2}}$.
Note that the linear transformation $x\mapsto\widetilde{x}$ keeps the Mahalanobis distances invariant, i.e., $\forall i,j\in [L]$, there is $\widetilde{\Delta}_{i,j}=\Delta_{i,j}$.
In the sequel, we will assume that the input pair $(x,y)$ satisfies Assumption~\ref{ass:3}.
For notation clarity, we denote $\widetilde{x}$ as $x$, $\widetilde{\mu}_i$ as $\mu_i$, $\widetilde{\Delta}_{i,j}$ as $\Delta_{i,j}$ without ambiguity.

This distribution is of interest as we can explicitly characterize the robustness to adversarial samples of a LDA classifier. Specifically,
the decision boundary obtained by LDA between class $i$ and $j$ is decided by the Fisher's linear discriminant function~\citep{friedman2001elements}, $\lambda_{i,j}(x)=\beta_{i,j}+\alpha_{i,j}^\top x=0$, where
\begin{equation*}\vspace{-0.cm}
\begin{split}
\beta_{i,j}&=\log(\pi_i/\pi_j)+\frac{1}{2}(\|\mu_j\|^2_2-\|\mu_i\|^2_2)\text{,}\\
\alpha_{i,j}^\top &=(\mu_i-\mu_j)^\top \text{.}
\end{split}
\end{equation*}
In the adversarial setting, the nearest adversarial example $x^*$ that satisfies condition~(\ref{adv}) w.r.t the normal example $x$ must be located on the decision boundary~\citep{Moosavidezfooli2016}.
We randomly sample a normal example of class $i$ as $x_{(i)}$, i.e., $x_{(i)}\sim\mathcal{N}(\mu_i,I)$, and denote its nearest adversarial counterpart on the decision boundary $\lambda_{i,j}(x)=0$ as $x_{(i,j)}^*$. According to condition~(\ref{adv}), there is $\hat{y}(x_{(i)})=i,\hat{y}(x_{(i,j)}^*)=j$ or $\hat{y}(x_{(i)})=j,\hat{y}(x_{(i,j)}^*)=i$, where $\hat{y}(\cdot)$ refers to the predicted label from the LDA classifier. We define the distance between $x_{(i)}$ and $x_{(i,j)}^*$ as $d_{(i,j)}$. Then we have the theorem on the relationship between the expectation $\ep[d_{(i,j)}]$ and the Mahalanobis distance $\Delta_{i,j}$:

\begin{theorem}
\label{the:2}
(Proof in Appendix A) If $\pi_i=\pi_j$, the expectation of the distance $d_{(i,j)}$ is a function of the Mahalanobis distance $\Delta_{i,j}$:
\begin{equation*}
\ep[d_{(i,j)}]=\sqrt{\frac{2}{\pi}}\exp\left( -\frac{\Delta_{i,j}^2}{8} \right) + \frac{1}{2}\Delta_{i,j}\left[ 1-2\Phi(-\frac{\Delta_{i,j}}{2}) \right]\text{,}
\end{equation*}
where $\Phi(\cdot)$ is the normal cumulative distribution function.
\end{theorem}

The more general result when $\pi_i\neq\pi_j$ can be found in the proof of Theorem~\ref{the:2}, which leads to similar conclusions. Furthermore, we can show that $\ep[d_{(i,j)}]$ monotonically increases w.r.t $\Delta_{i,j}$, as summarized in the following corollary.

\begin{corollary}\label{co:1}
The partial derivative of $\ep[d_{(i,j)}]$ w.r.t $\Delta_{i,j}$ is
\begin{equation*}
\frac{\partial \ep[d_{(i,j)}]}{\partial \Delta_{i,j}}=\frac{1}{2}[1-2\Phi(-\frac{\Delta_{i,j}}{2})]\geq0\text{,}
\end{equation*}
where the Mahalanobis distance $\Delta_{i,j}$ is non-negative.
\end{corollary}

\citet{Moosavidezfooli2016} define the robustness of a point $x_{(i)}$ as $\min_{j\neq i}d_{(i,j)}$. Similar to this definition, we define the robustness of the classifier as below. Note that $\ep[d_{(i,j)}]$ is the expectation value of the minimal distance from a normal example to its potential adversarial counterpart between class $i$ and $j$. Thus $\ep[d_{(i,j)}]$ can measure the local robustness of the classifier on the attacks focusing on the two classes, where a larger value of $\ep[d_{(i,j)}]$ indicates better local robustness, and vice verse. Then the robustness of the classifier on all the attacks can be measured by
\begin{equation}
\RB=\min_{i,j\in[L]}\ep[d_{(i,j)}]\text{,}
\end{equation}
which can be regarded as a tight lower bound of the local robustness between any two classes. Because $\ep[d_{(i,j)}]$ monotonically increases w.r.t $\Delta_{i,j}$, we prefer larger values of $\Delta_{i,j}$ for better local robustness. According to Corollary~\ref{co:1}, the gap $\left|\ep[d_{(i,j)}]/\Delta_{i,j}-1/2\right|$ monotonically decreases to $0$ w.r.t $\Delta_{i,j}$, e.g., when $\Delta_{i,j}>10$, we can numerically figure out that $\left|\ep[d_{(i,j)}]/\Delta_{i,j}-1/2\right|<10^{-7}$. Thus we can approximate $\ep[d_{(i,j)}]$ using $\Delta_{i,j}/2$, which further results in an approximation for the robustness $\RB$ as
\begin{equation}
\RB\approx\overline{\RB}=\min_{i,j\in[L]}\Delta_{i,j}/2\text{.}
\end{equation}
We now investigate when the approximated robustness $\overline{\RB}$ of the LDA classifier can achieve its maximal value, and derive an efficient algorithm to estimate the unknown means $\mu=\{\mu_i|i\in[L]\}$ of the input distribution.
Let $\left\|\mu \right\|_2$ be $\max_i\left\|\mu_i \right\|_2$. Since $\mu$ has finite elements, there always exists a positive constant $C$, such that $\left\|\mu \right\|_2^2=C$. The following theorem gives a tight upper bound for $\overline{\RB}$.


\begin{theorem}
(Proof in Appendix A) Assume that $\sum_{i=1}^L\mu_i=0$ and $\left\|\mu \right\|_2^2=C$. Then we have
\begin{equation*}
\overline{\RB}\leq \sqrt{\frac{LC}{2(L-1)}}\text{.}
\end{equation*}
The equality holds if and only if
\begin{equation}\label{eq:4}
\mu_i^\top \mu_j=\begin{cases}
C\text{,}& i=j\text{,}\\
C/(1-L)\text{,}& i\neq j\text{,}
\end{cases}
\end{equation}
where $i,j\in[L]$ and $\mu_i,\mu_j \in \mu$.
\end{theorem}

\begin{algorithm}[t]
\caption{GenerateOptMeans}
\label{algo:1}
\begin{algorithmic}
\STATE {\bfseries Input:} The constant $C$, the dimension of vectors $p$ and the number of classes $L$. ($L\leq p+1$)\\
\STATE {\bfseries Initialization:} Let the $L$ mean vectors be $\mu_1^*=e_1$ and $\mu_i^*=0_p,i\neq 1$. Here $e_1$ and $0_p$ separately denote the first unit basis vector and the zero vector in $\R^p$.
\FOR{$i=2$ {\bfseries to} $L$}
\FOR{$j=1$ {\bfseries to} $i-1$}
\STATE $\mu^*_i(j)=-[1+\langle \mu^*_i,\mu^*_j \rangle \cdot (L-1)]/[\mu^*_j(j) \cdot(L-1)]$
\ENDFOR
\STATE $\mu^*_i(i)=\sqrt{1-\lVert\mu^*_i \rVert^2_2}$
\ENDFOR
 \FOR{$k=1$ {\bfseries to} $L$}
\STATE $\mu^*_k=\sqrt{C}\cdot \mu^*_k$
\ENDFOR
\STATE {\bfseries Return:} The optimal mean vectors $\mu^*_i,i\in[L]$.
\end{algorithmic}
\end{algorithm}

We denote any set of means that satisfy the optimal condition~(\ref{eq:4}) as $\mu^*$. When $L\leq p+1$,\footnote{Otherwise, there is no solution for $\mu^*$.}
there is an infinite number of $\mu^*$ because of the degeneracy of the condition.
Intuitively, the elements in $\mu^*$ constitute the vertexes of an equilateral triangle when $L=3$, and those of a regular tetrahedron when $L=4$. In Alg.~\ref{algo:1}, we propose an easy-to-implement method to construct a set of means $\mu^*_0$ that satisfy the condition, where $\mu^*_0=\GOM(C,p,L)$.

With the above results, we formally define a joint distribution with the form
\begin{equation*}
P(y=i)=\pi_{i}\text{, }P(x|y=i)=\mathcal{N}(\mu^*_{i},\I)\text{, }i\in[L]
\end{equation*}
as a \emph{Max-Mahalanobis distribution (MMD)}, namely, it has the maximal minimal Mahalanobis distance between any two Gaussian components for a given $\left\|\mu \right\|_2$. In a nutshell, when regarding the set of means $\mu$ in Assumption~\ref{ass:3} as independent variables for a given $\left\|\mu \right\|_2$, the LDA classifier would have the best robustness if its input distributes as a MMD.
We refer to the above process as the \emph{Max-Mahalanobis linear discriminant analysis (MM-LDA)} procedure.

\subsection{The MM-LDA Network}\label{MMLDA}
Though elegant, the MM-LDA procedure is not directly applicable in practice, as the mixture of Gaussian Assumption~\ref{ass:3} is unlikely to hold in the input space (e.g., images in the pixel space), in which the data distribution $P(x)$ can be very complex.
Fortunately, thanks to the universal approximation power of neural networks~\citep{Hornik1989} and the algorithmic advances, previous work on deep generative models~\cite{goodfellow2014generative,kingma2013auto} has demonstrated that a deep neural network can be learned to transform a simple distribution (e.g., standard normal) to a complex one that matches the data distribution.
The reverse direction is also true, and it has been implicitly applied in feed-forward networks, where a powerful nonlinear transformation network is learned to turn a complexly distributed input data into a latent feature space, and then a linear classifier (e.g., SR) is sufficient to achieve state-of-the-art performance~\citep{simonyan2014very,He2015,He2016,szegedy2016rethinking}.
Therefore, we can expect that the MM-LDA procedure will work well on a properly learned latent feature space by exploring the power for DNNs, as detailed below.

Formally, we propose the \emph{Max-Mahalanobis linear discriminant analysis (MM-LDA)} network, which consists of a nonlinear transformation network (characterized as a DNN) to turn the input $x$ into a latent feature representation $z$, and applies the MM-LDA procedure on $z$. Namely,
the MM-LDA network explicitly models $P(z)$ as a MMD, and applies LDA on $z$ to make predictions.
According to the analysis in Section~\ref{MMD}, the MM-LDA network can have the best robustness in the latent feature space, and further results in better robustness in the input space. 


\begin{algorithm}[t]
\caption{The training phase for the MM-LDA network}
\label{algo:2}
\begin{algorithmic}
\STATE {\bfseries Input:} The model $z_{\theta}(x)$, the square norm $C$ of Gaussian means, the training dataset $\mathcal{D}=\{(x_i, y_i)\}_{i\in [N]}$.\\
\STATE {\bfseries Initialization:} Initialize $\theta$ as $\theta_0$, the training step as $s=0$. Let $p=\dim(z)$, $\varepsilon$ be the learning rate variable.
\STATE Get $\mu^*=\GOM(C,p,L)$ for the MMD.
\WHILE{not converged}
\STATE Sample a mini-batch of training data $\mathcal{D}_m$ from $\mathcal{D}$,
\STATE Calculate the objective
\[\mathcal{L}_{\text{MM}}^m=\frac{1}{\left|\mathcal{D}_m\right|} \sum_{(x_i, y_i) \in \mathcal{D}_m}\mathcal{L}_{\text{MM}}(x_i, y_i,\mu^*)\text{,}\]
\STATE Update parameters $\theta_{s+1} \leftarrow\theta_s-\varepsilon\nabla_{\theta}\mathcal{L}_{\text{MM}}^m$,
\STATE Set $s\leftarrow s+1$.
\ENDWHILE
\STATE {\bfseries Return:} The parameters $\theta_{\text{MM}}=\theta_s$.
\end{algorithmic}
\end{algorithm}

Given a feature vector $z$ in the MM-LDA network, according to Bayes' theorem and the definition of MMD, we have the conditional distribution of labels:
\begin{equation*}
P(y\!=\!k|z)=\frac{P(z|y\!=\!k)P(y\!=\!k)}{P(z)}=\frac{\pi_{k}\mathcal{N}(z|\mu^*_{k},\I)}{\sum_{i=1}^L\pi_{i}\mathcal{N}(z|\mu^*_{i},\I)}\text{.}
\end{equation*}
Note that the feature vector $z$ is actually $z_{\theta}(x)$, since it is obtained by the nonlinear transformation network $x\mapsto z$, parameterized by $\theta$. 
Instead of estimating model parameters from data as LDA does, MM-LDA treats the set of means $\mu^*$ and different class priors $\pi_i,i\in [L]$ as hyperparameters, and $\theta$ is what the MM-LDA network needs to learn in the training phase.
Similar to the SR network, we let $F_{\text{MM}}(x)$ be the output prediction of the MM-LDA network. The $k$-th element of the prediction is
\begin{equation}\label{eq:5}
F_{\text{MM}}(x)_k=P(y=k|z_{\theta}(x))\text{,}
\end{equation}
where $k\in[L]$. In the training phase, the loss function\footnote{More discussion on the training loss function for the MM-LDA network can be found in Appendix B.1} for the MM-LDA network is $\mathcal{L}_{\text{MM}}(x,y)=-1_y^\top \log F_{\text{MM}}(x)$,
which is the cross-entropy between the one-hot true label $1_y$ and the prediction $F_{\text{MM}}(x)$. By minimizing the loss function w.r.t $\theta$ on the training set $\mathcal{D}:=\{(x_i,y_i)\}_{i\in[N]}$, we can obtain the optimal parameters $\theta_{\text{MM}}$ for the MM-LDA network as $\theta_{\text{MM}}=\argmin_{\theta}\frac{1}{N}\sum_{i=1}^{N}\mathcal{L}_{\text{MM}}(x_i,y_i)$.
In Alg.~\ref{algo:2} we demonstrate the complete training phase. In the test phase, the MM-LDA network returns the predicted label $\hat{y}_{\text{MM}}=\argmax_{k}F_{\text{MM}}(x)_{k}$, where the set of means $\mu^*$ is the same as the one used in training.

Fig.~\ref{fig:2} provides an intuitive comparison between MM-LDA networks and SR networks (See Sec.~\ref{PNE} for details). For SR networks, though the latent features are discriminative, the distribution is not as orderly as that for MM-LDA networks. This is because SR networks do not explicitly model the distribution of $z$, while MM-LDA does model it as a well-structured MMD. This structure can influence the nonlinear transformation network via back-propagation as in Alg.~\ref{algo:2}.

In addition to better robustness in the adversarial setting, the MM-LDA network should also perform better than the SR network on the input with a class-biased distribution, i.e., a distribution with different class priors. This can be intuitively illustrated by the conclusions in Section~\ref{ARE} that a larger value of $\left|\zeta\right|$ implies lower relative efficiency of LR to LDA. Since $\zeta$ denotes the log-ratio of class-priors, a larger value of $\left|\zeta\right|$ indicates more biased class priors, i.e., bigger gaps among class priors. Thus in the multi-class cases, acting on the input with biased class priors should intuitively result in low relative efficiency of SR to LDA, and further to MM-LDA.




\vspace{-.1cm}
\section{Experiments}
\vspace{-.1cm}
We now experimentally demonstrate that the MM-LDA networks are more robust in the adversarial setting while maintaining state-of-the-art performance on normal examples, and have better performance on class-biased datasets.

\vspace{-.1cm}
\subsection{Setup}
\vspace{-.1cm}
We choose the widely used MNIST~\citep{Lecun1998} and CIFAR-10~\citep{Krizhevsky2012} datasets. MNIST consists of grey images of handwritten digits in classes $0$ to $9$, and CIFAR-10 consists of color images in $10$ different classes. Each dataset has 60,000 images, of which 50,000 are in the training set and the rest are in the test set. The pixel values of images in both sets are scaled to the interval $[-0.5, 0.5]$ before fed into classifiers. The baseline is the most common SR network~\citep{Goodfellow-et-al2016}. The empirical class prior $\hat{\pi}_k$ of a dataset with $N$ samples is $\pi_k = N_k / N$, where $N_k$ is the number of samples in class $k$. 
Then a dataset is class-unbiased if $\forall k\in[L],\hat{\pi}_k=1/L$ for both training and testing sets; otherwise class-biased. 

\vspace{-.1cm}
\subsection{Performance on Normal Examples}\label{PNE}
\vspace{-.1cm}

\begin{figure}[t]
\vspace{-.18cm}
\setlength{\fboxrule}{0.3pt}
\centering
\subfloat[Resnet-32 (SR)]{\fbox{\includegraphics[width = 0.45\columnwidth]{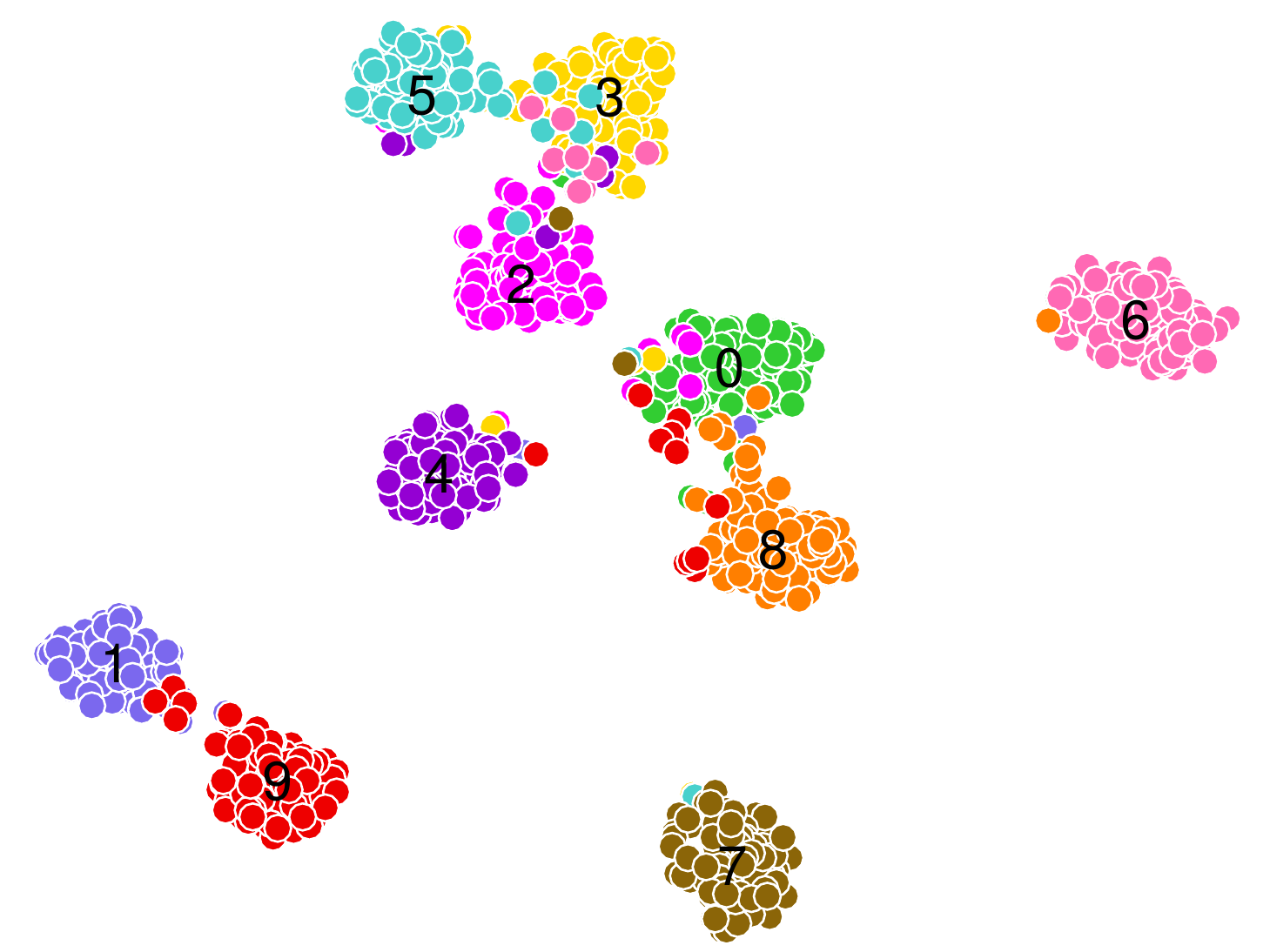}}\label{fig:2a}}\
\hspace{-0.01\linewidth}
\subfloat[Resnet-32 (MM-LDA)]{\fbox{\includegraphics[width = 0.45\columnwidth]{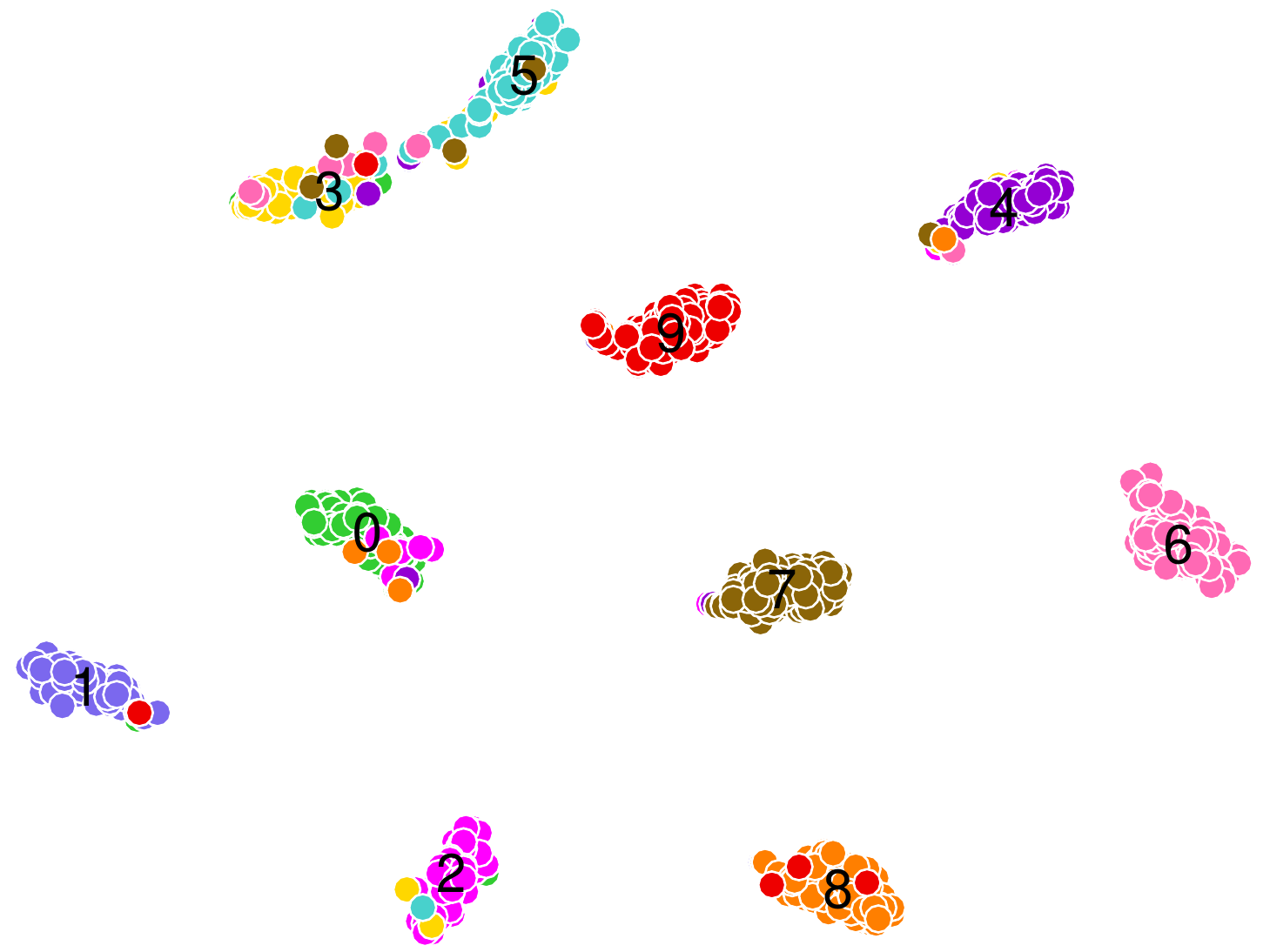}}\label{fig:2b}}
\vspace{-.15cm}
\caption{t-SNE visualization of the latent features on CIFAR-10. The index numbers indicate classes $0$ to $9$, where each number locates on the median position of the corresponding vectors.}
\label{fig:2}
\vspace{-.4cm}
\end{figure}

\begin{table*}[t]\vspace{-.55cm}
  \caption{Classification accuracy (\%) on adversarial examples of MNIST and CIFAR-10. The investigated values of perturbation are 0.04, 0.12, and 0.20. \textbf{Boldface} indicates the best result under certain combination of a value of perturbation and an attacking method.}
  \label{table2}
  \begin{center}
  \begin{small}
  \begin{tabular}{|c|c|c|c|c|c|c|c|c|c|}
   \hline
     \multirow{2}{*}{Perturbation}& \multirow{2}{*}{Model} & \multicolumn{4}{c|}{\textbf{MNIST}} &\multicolumn{4}{c|}{\textbf{CIFAR-10}}  \\
     & & FGSM & BIM & ILCM & JSMA & FGSM & BIM & ILCM & JSMA \\
    \hline
    \hline
    \multirow{4}{*}{0.04}&Resnet-32 (SR) &93.6  &87.9 &94.8&92.9&20.0&5.5&0.2&65.6 \\
               &Resnet-32 (SR)\ +\ SAT &86.7  &68.5 &98.4&   -      &   24.4  &  7.0 & 0.4  & -    \\
               &Resnet-32 (SR)\ +\ HAT &88.7  &96.3 &\textbf{99.8}&   -      &30.3&5.3&1.3& -    \\
    &Resnet-32 (MM-LDA) & \textbf{99.2}  & \textbf{99.2} &99.0& \textbf{99.1}& \textbf{91.3}& \textbf{91.2}& \textbf{70.0}& \textbf{91.2}\\
    \hline
    \multirow{4}{*}{0.12}&Resnet-32 (SR)  &28.1  &3.4 &20.9&56.0&10.2&4.1&0.3&20.5 \\
                     &Resnet-32 (SR)\ +\ SAT &40.5  &8.7 &88.8&   -      &  88.2 &  6.9 & 0.1   & -    \\
                &Resnet-32 (SR)\ +\ HAT &40.3  &40.1 &92.6&   -    &44.1&8.7&0.0& -   \\
    &Resnet-32 (MM-LDA) & \textbf{99.3}  & \textbf{98.6} & \textbf{99.6}& \textbf{99.7}& \textbf{90.7}& \textbf{90.1}& \textbf{42.5}& \textbf{91.1}\\
    \hline
    \multirow{4}{*}{0.20}&Resnet-32 (SR)  &15.5  &0.3 &1.7&25.6&10.7&4.2&0.6&11.5 \\
                   &Resnet-32 (SR)\ +\ SAT &17.3  &  1.1   &69.4&   -      &  \textbf{91.7}  &  9.4  &  0.0  & -    \\
                &Resnet-32 (SR)\ +\ HAT &10.1  &10.5 &46.1&    -    &40.7&6.0&0.2&  -  \\
    &Resnet-32 (MM-LDA) & \textbf{97.5}  & \textbf{97.3} & \textbf{96.6}& \textbf{99.6}& 89.5& \textbf{89.7}& \textbf{31.2}& \textbf{91.8}\\
\hline
      \end{tabular}
  \end{small}
  \end{center}
\vspace{-.7cm}
\end{table*}

We first test the performance on normal examples (i.e., the samples in the original datasets without any perturbations).
We implement Resnet-32~\citep{He2015} on MNIST and CIFAR-10, which uses the SR model as the linear classifier. This network will be denoted by Resnet-32 (SR).
Our MM-LDA network shares the same architecture of nonlinear transformation as Resnet-32 (SR) while uses the MM-LDA procedure for classification, and we denote it by Resnet-32 (MM-LDA).
The number of training steps is 20,000 on MNIST and 90,000 on CIFAR-10 for both networks. Here we apply the training setting introduced in \citet{He2016} to train the Resnet-32 (SR). To train the Resnet-32 (MM-LDA), we simply use the same training setting as the Resnet-32 (SR), except that we apply the adaptive optimization method---Adam~\citep{Kingma2014} rather than the momentum SGD used in \citet{He2016}, to avoid extra effort on tuning training hyperparameters.\footnote{We also try to substitute Adam for the momentum SGD in the training phase of Resnet-32 (SR), and we found that the momentum SGD makes Resnet-32 (SR) have lower error rates on both datasets.}

\begin{wrapfigure}{r}{0.23\textwidth}
\vspace{-0.28in}
\begin{center}
\includegraphics[width=0.23\textwidth]{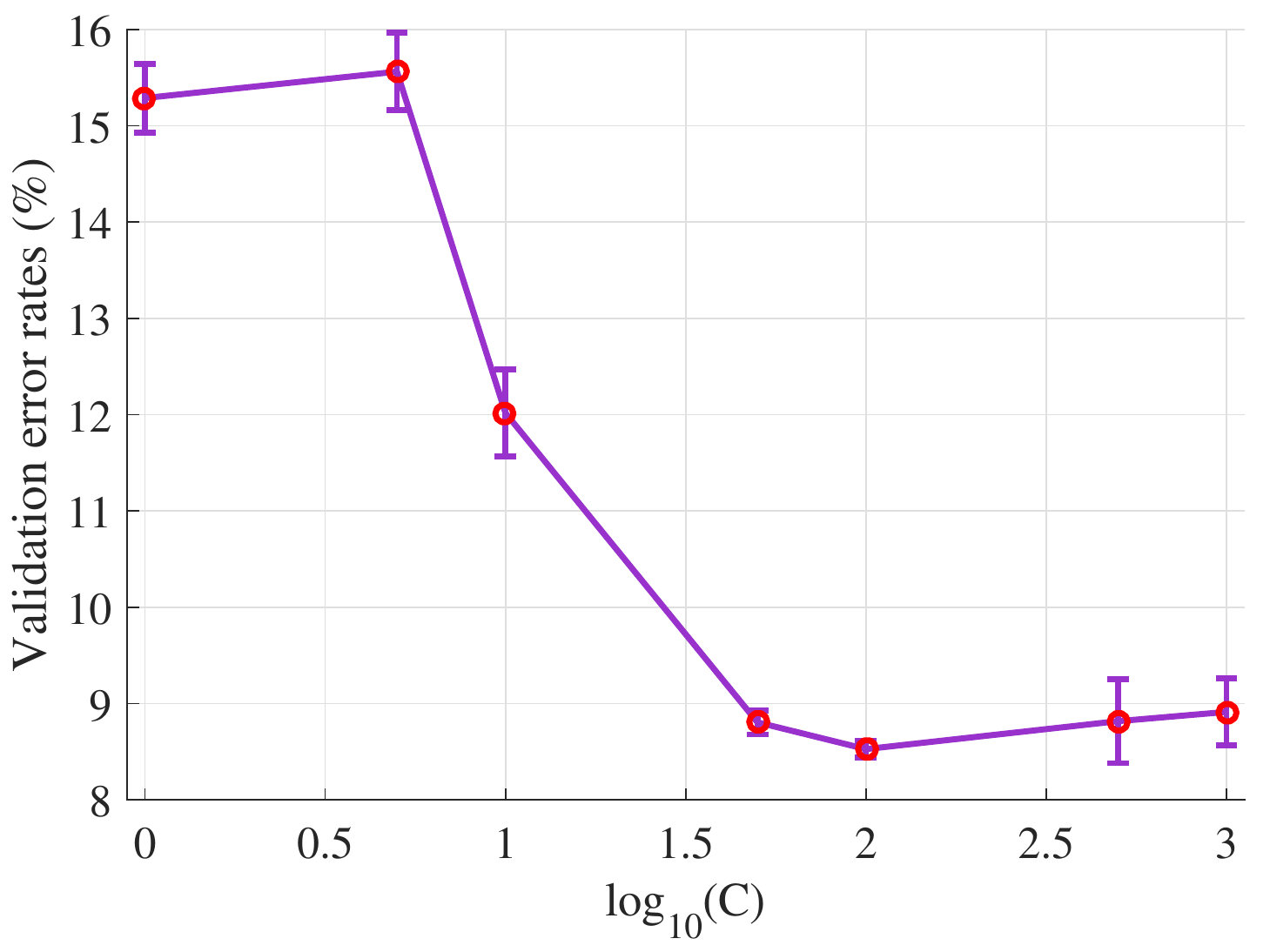}
\vspace{-.8cm}
\caption{Validation error rates (\%) of the MM-LDA networks w.r.t different values of $\log_{10}(C)$ on CIFAR-10.}
\label{fig:5}
\end{center}
\vspace{-0.05in}
\end{wrapfigure}

When applying the MM-LDA network, the only hyperparameter is the square norm $C$ of the Gaussian means in MMD. If $C$ is too small, the conditional Gaussian components in MMD will largely overlap to each other, which makes the optimal Bayes error rate be high, and further results in a high error rate for the LDA classifier. Besides, if $C$ is too large, the magnitudes of the transformation parameters $\theta$ will also tend to sharply increase during the training procedure, which makes the MM-LDA network easy to overfit. In our experiments, we empirically choose the value of $C$ by doing $5$-fold cross-validation on the training set. 
Fig.~\ref{fig:5} shows the validation error rates of the MM-LDA networks w.r.t $\log_{10}(C)$ on CIFAR-10. We find that when $\log_{10}(C)=2$, i.e., $C=100$ the MM-LDA network has the lowest average validation error rate with a small value of standard deviation. Thus hereafter we will set $C=100$.




Table~\ref{table1} shows the classification error rates on the test sets of both datasets. We can see that the MM-LDA network maintains state-of-the-art performance on normal examples. Furthermore, in Fig.~\ref{fig:2} we apply the t-SNE technique~\citep{Hinton2008} to visualize the latent feature vectors on 1,000 randomly sampled test images of CIFAR-10. We can find that the trained MM-LDA network maps the data distribution in the input space to a much more regular distribution in the latent feature space, as stated before. Note that class $3$ and class $5$ are close to each other in Fig.~\ref{fig:2b}, which is reasonable since they are separately 'cat' and `dog', even a human observer will sometimes misidentify them.

\vspace{-.1cm}
\subsection{Performance in the Adversarial Setting}
\vspace{-.1cm}
Now we test the robustness of MM-LDA networks in the adversarial setting. Adversarial training is one of the most common and effective methods to improve the robustness of classifiers on iterative-based attacks~\citep{Szegedy2013,Goodfellow2014,kurakin2016adversarial,tramer2017ensemble}. Thus in addition to the SR networks trained on normal examples, we also treat the SR networks enhanced by adversarial training as stronger baselines. We construct the enhanced baselines by first crafting adversarial examples on the trained SR networks, then fine-tuning the networks on the mixture of the normal examples and crafted adversarial examples.
More technical details are in Appendix B.2.

For more complete analysis, we apply two kinds of adversarial training methods to enhance baselines. They differ in the choices of adversarial examples to fine-tune the classifiers:

\textbf{Specific Adversarial Training (SAT)} fine-tunes the classifiers on the adversarial examples crafted by the same attack with the same value of perturbation $\epsilon$ as that when attacking the classifiers.
Similar strategy is used in \citep{Szegedy2013,Goodfellow2014}.

\textbf{Hybrid Adversarial Training (HAT)} fine-tunes the classifiers on the adversarial examples crafted by the same attack as that when attacking the classifiers, but with various values of $\epsilon$. Specifically, we uniformly choose $\epsilon$ from the interval $[0.02,0.20]$ when crafting adversarial examples for HAT. Similar strategy is used in \citep{kurakin2016adversarial}.

\begin{table}[t]       \vspace{-0.1in}
  \caption{Error rates (\%) on the test sets of MNIST and CIFAR-10.}
  \label{table1}
  \begin{center}
  \begin{small}
  \begin{tabular}{|c|c|c|}
   \hline
     Model & MNIST & CIFAR-10  \\
    \hline
    \hline
    Resnet-32 (SR) &0.38  & \textbf{7.13} \\
    Resnet-32 (MM-LDA) & \textbf{0.35}  & 8.04 \\
    \hline
      \end{tabular}
  \end{small}
  \end{center}
    \vspace{-0.3in}
\end{table}

Table~\ref{table2} presents the classification accuracy of the networks on the adversarial examples crafted by iterative-based attacks. We investigate on three different values of perturbation $\epsilon$: 0.04, 0.12 and 0.20 (See Section~\ref{AS} for $\epsilon$). 
Usually an adversarial noise with perturbation larger than $0.05$ is perceivable by human eyes.
From the results, we can see that both SAT and HAT can effectively enhance the original baselines to stronger ones. However, the adversarial training methods require extra computational cost and are less effective on multi-step methods, e.g., BIM and ILCM~\citep{kurakin2016adversarial}. By contrast, the MM-LDA network significantly improves the robustness on iterative-based attacks compared to almost all the baselines. This is because in the MM-LDA networks, normal examples distribute as a MMD in the latent feature space, which makes it more difficult for adversaries to move a normal example from its original class to other classes. Note that we do not adversarially fine-tune Resnet-32 (SR) on the JSMA attack, since it is computationally expensive to craft an adversarial example by JSMA, which makes adversarial training inefficient.

We also apply the optimization-based C\&W attack on the trained networks. Since there is yet no method including adversarial training to effectively defend the C\&W attack under the white-box threat model~\citep{Carlini2016,carlini2017adversarial}, we only compare between Resnet-32 (SR) and Resnet-32 (MM-LDA).  In the C\&W attack, we set the binary search steps for the constant $c$ be $9$, and the maximal number of iteration steps for each value of $c$ be 10,000. This setting is strong enough, so that the crafted adversarial examples can evade both the SR and MM-LDA networks with nearly 100\% success rate. As shown in Table~\ref{table3}, the average minimal distortions of adversarial examples on the MM-LDA networks are much larger than those on the SR networks. Here the distortion is defined in~\citet{Szegedy2013}, where the pixel values of images are in the interval $[0,255]$ when calculating it. This results mean that the C\&W attack has to add much larger noises to successfully evade the MM-LDA networks, as theoretically demonstrated in Sec.~\ref{MMD}.

\begin{figure*}[t]\vspace{-.28cm}
\begin{center}
\includegraphics[width=2\columnwidth]{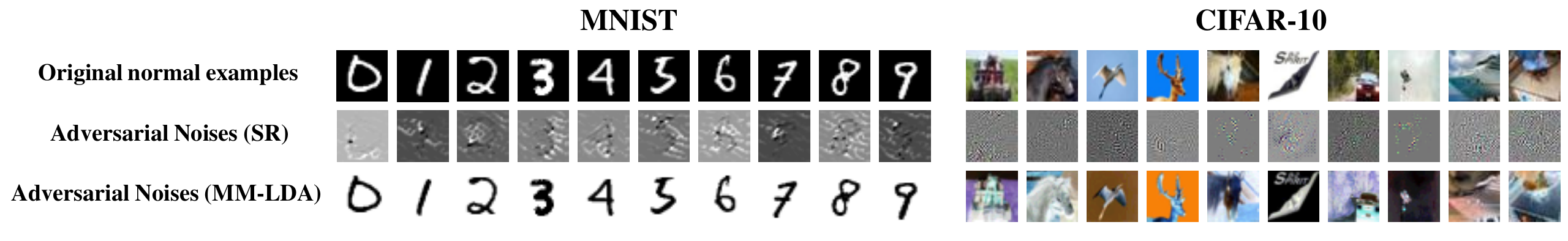}
\vspace{-.3cm}
\caption{Some normal examples with the semantic adversarial noises crafted by the C\&W attack on MNIST and CIFAR-10. The shown noises crafted for the MM-LDA networks are much more semantic, while similar meaningful noises are hardly observed in those crafted for the SR networks. \emph{However, most of the adversarial noises crafted for the MM-LDA networks still seem like random noises.}}
\label{fig:4}
\end{center}
\vspace{-.35cm}
\end{figure*}

\begin{table}[t]  \vspace{-0.1in}
  \caption{Average minimal distortions of the adversarial examples crafted by the C\&W attack on MNIST and CIFAR-10.}
  \label{table3}
  \begin{center}
  \begin{small}
  \begin{tabular}{|c|c|c|c|}
   \hline
     Model & MNIST & CIFAR-10  \\
    \hline
    \hline
    Resnet-32 (SR)  &8.56  &0.67  \\
    Resnet-32 (MM-LDA)  & \textbf{16.32}  & \textbf{2.80} \\
    \hline
      \end{tabular}
  \end{small}
  \end{center}
  \vspace{-0.3in}
\end{table}

Furthermore, we find that when applying the C\&W attack on the MM-LDA networks, some of the adversarial noises have the same semantic meanings as their corresponding normal examples (around $1\%\sim5\%$ of all noises), while similar phenomenon can hardly be observed when attacking on the SR networks (less than $0.1\%$ of all noise). We show some of the semantic noises in Fig.~\ref{fig:4}. The adversarial noise is calculated as $(x^*-x)/2$ to keep the pixel values of the noise in the interval $[-0.5,0.5]$. This result indicates that MM-LDA networks can learn more robust features,
such that on the shown normal examples, the optimal attacking strategy that the C\&W attack finds for MM-LDA networks is to weaken the features of the normal examples as a whole, rather than adding meaningless noise as for the SR networks.

\vspace{-.1cm}
\subsection{Performance on Class-biased Datasets}
\vspace{-.1cm}
Finally, we evaluate on class-biased datasets, which are more realistic though many sets were artificially constructed as class-unbiased, e.g., CIFAR-10. 
We construct the class-biased datasets by randomly sampling each data point of class $i$ from CIFAR-10 with probability $\alpha_i,i\in L$, where $L=10$ for CIFAR-10. Specifically, let $\alpha=(\alpha_0, \cdots,\alpha_{9})$, $\mathcal{D}$ be the training or test set of CIFAR-10, then the constructed class-biased dataset is $\mathcal{D}^{\text{bias}}_{\alpha}$ that $\forall (x,y)\in\mathcal{D}$, $P((x,y)\in\mathcal{D}^{\text{bias}}_{\alpha})=\alpha_y$. Then the empirical class priors of $\mathcal{D}^{\text{bias}}_{\alpha}$ have expectations as $\ep[\hat{\pi}_k]=\alpha_k/\|\alpha\|_1$. We choose two typical kinds of bias probability $\alpha$ as below:

\textbf{Bias Probability 1 (BP1)} has $\alpha=(0.1,0.2,0.3,\cdots,1.0)$. To avoid the system error caused by certain  permutation of the elements in $\alpha$, we randomly rearrange the elements in $\alpha$ to get 10 counterparts $\alpha^{(0)},\cdots\alpha^{(9)}$. The publicly available datasets with similar class-prior distributions as BP1 including the IMDB-WIKI dataset~\citep{rothe2015dex} for age and gender prediction, and the KITTI dataset~\citep{Geiger2012CVPR} for autonomous driving.

\textbf{Bias Probability 2 (BP2)} has $\alpha=(0.2,\cdots,0.2,1.0)$. Since there is only one element in $\alpha$ that equals to $1.0$ and the others all equal to $0.2$, we assign $1.0$ in turn to 10 different classes to avoid the system error, and similarly get 10 counterparts $\alpha^{(0)},\cdots,\alpha^{(9)}$. The Caltech101 dataset~\citep{fei2007learning} and the large-scale ImageNet dataset~\citep{deng2009imagenet} have similar class-prior distributions as BP2.

\begin{figure}[t] \vspace{-0.12in}
\centering
\subfloat[BP1]{\includegraphics[width = 0.49\columnwidth]{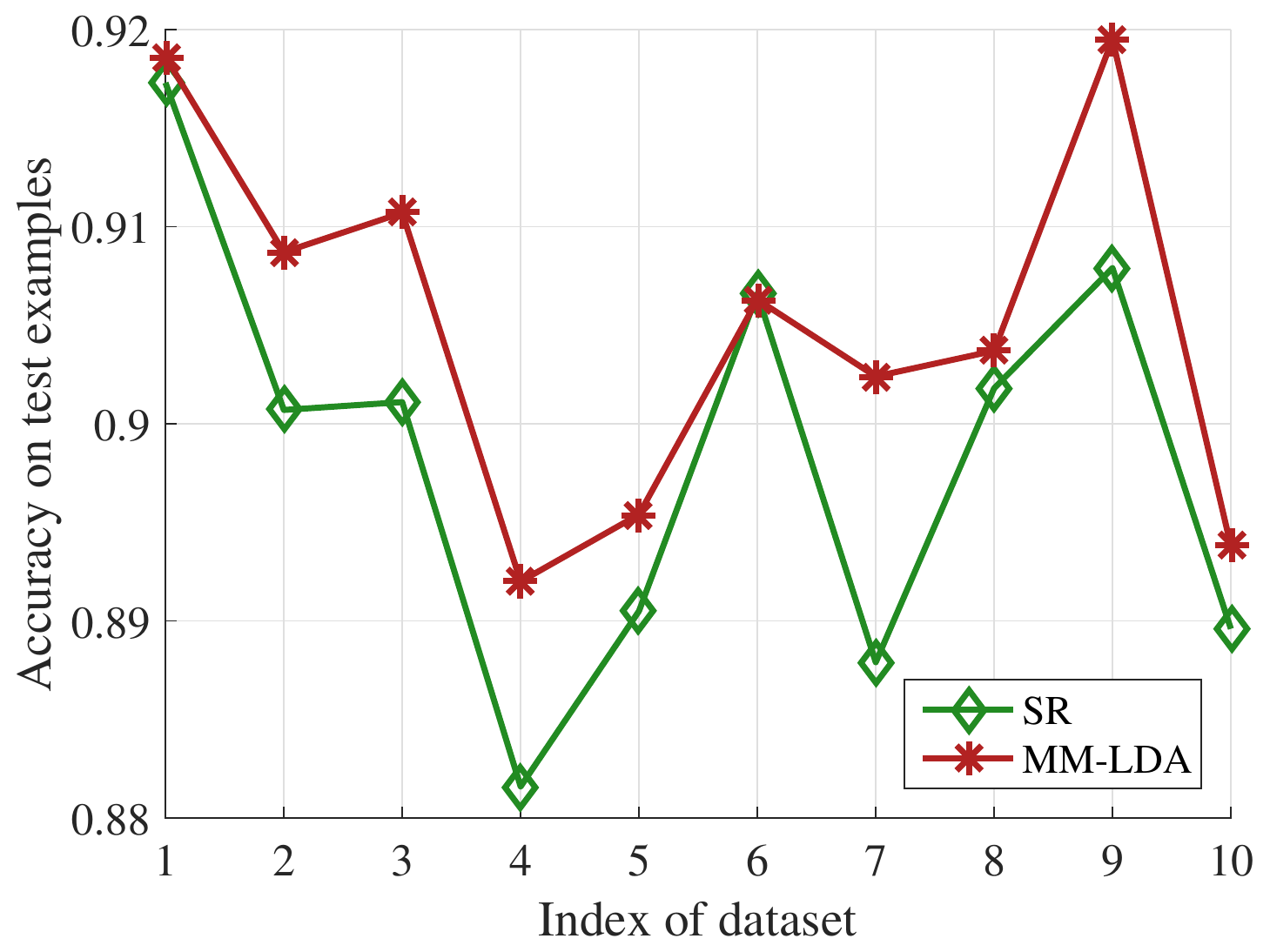}}\
\hspace{-0.01\linewidth}
\subfloat[BP2]{\includegraphics[width = 0.49\columnwidth]{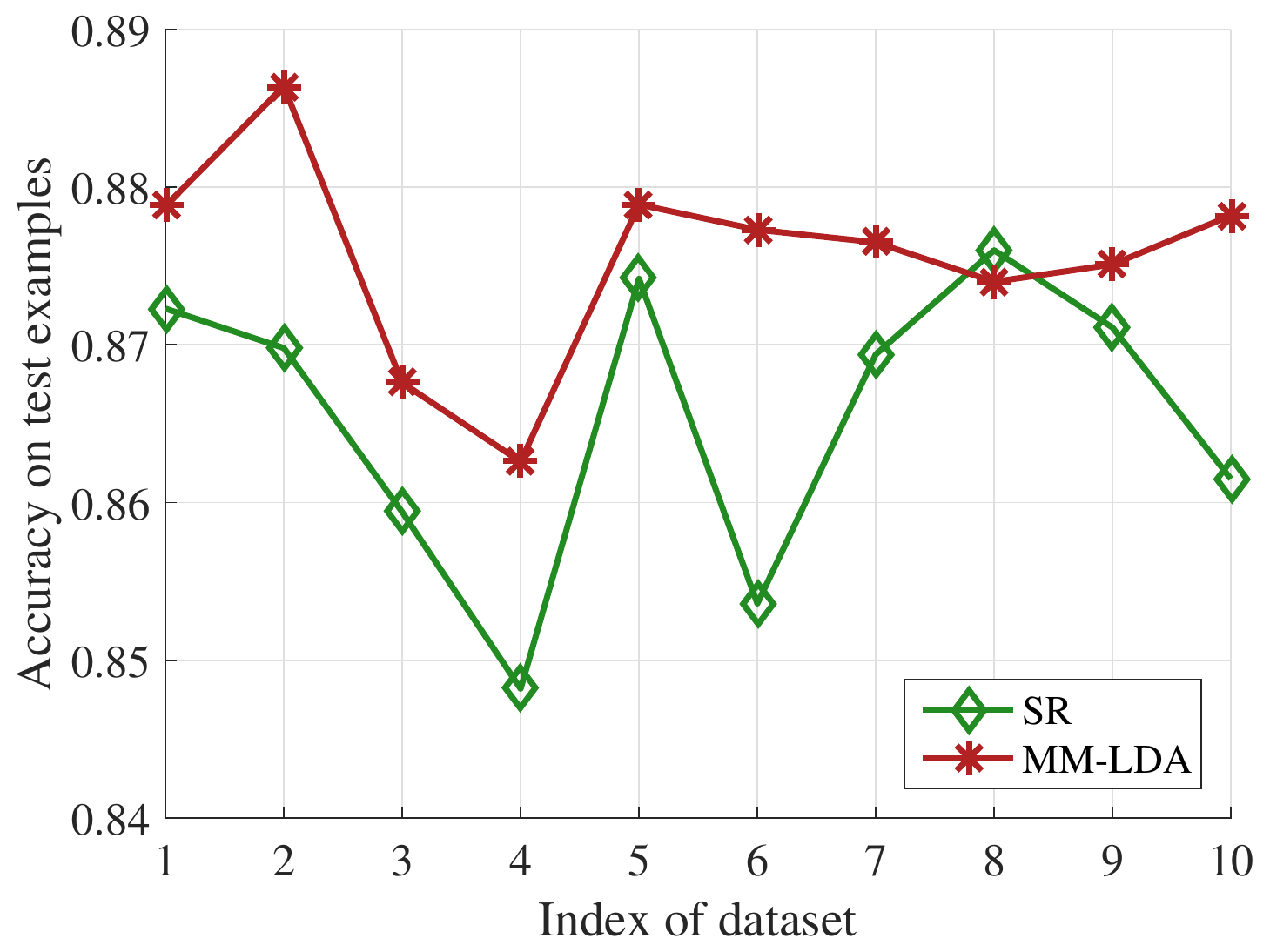}}
\vspace{-.3cm}
\caption{Classification accuracy on the test sets of class-biased datasets. Each index of dataset corresponds to a counterpart of the bias probability. The original class-unbiased dataset is CIFAR-10.}
\label{fig:3}\
 \vspace{-0.43in}
\end{figure}

We separately apply the counterparts of BP1 and BP2 on both the training and test sets of CIFAR-10 to construct totally 20 class-biased datasets, i.e., both the training and test sets of each constructed dataset are class-biased. Then we train Resnet-32 (SR) and Resnet-32 (MM-LDA) on each class-biased dataset. Fig.~\ref{fig:3} shows the test accuracy of the trained networks on the 20 class-biased datasets. Note that when training and testing the MM-LDA networks, we set the class priors as uniform $\pi_k=1/L$ to calculate the prediction $F_{\text{MM}}(x)$ rather than setting $\pi_k=\alpha_k/\|\alpha\|_1$. By doing this we give a fair comparison between the SR and the MM-LDA networks, since SR is a discriminant model that cannot exploit the information of class priors. We can see that the MM-LDA networks still perform better than the SR networks on almost all the datasets constructed by the counterparts of BP1 and BP2. This result indicates that the  better performance of the MM-LDA networks on class-biased datasets comes from the intrinsic superiority of the MM-LDA networks, not from the extra knowledge on class priors. We also try to set $\pi_k=\alpha_k/\|\alpha\|_1$, and find that the difference of $F_{\text{MM}}(x)$ between the two settings is small, which most likely leads to the same predicted label. This is because when the class priors are not too biased, i.e., the value of $\max_{i,j\in[L]}\left|\log(\pi_i/\pi_j)\right|$ is not too large, the exponential terms in Eq.~(\ref{eq:5}) will dominate the calculation of $F_{\text{MM}}(x)$ since we choose a relatively large $C=100$.


\vspace{-.1cm}
\section{Conclusions}
\vspace{-.1cm}
In this paper we propose the novel MM-LDA network. The MM-LDA network is much more robust in the adversarial setting with theoretical guarantees, while maintaining state-of-the-art performance on normal examples. The MM-LDA network also performs better on class-biased datasets. Our network is easy to implement and can be naturally combined with different nonlinear transformation architectures and training methods designed for the SR network.

\newpage
\section*{Acknowledgements}
This work was supported by NSFC Projects (Nos. 61620106010, 61621136008, 61332007), Beijing NSF Project (No. L172037), Tiangong Institute for Intelligent Computing, NVIDIA NVAIL Program, Siemens and Intel.


\bibliography{reference}
\bibliographystyle{icml2018}

 \clearpage
 \appendix
\section{Proof}
\begin{theoremappendix}
The relative efficiency of logistic regression to linear discriminant analysis is

\begin{equation*}
\eff_p(\zeta,\Delta)=(Q_1+(p-1)Q_2)/(Q_3+(p-1)Q_4)\text{,}
\end{equation*}

where $Q_2=1+\pi_0\pi_1\Delta^2$, $Q_4=\frac{1}{A_0}$ and

\begin{equation*}
\begin{split}
Q_1&=\begin{pmatrix}
    1 & \frac{\zeta}{\Delta}
  \end{pmatrix}
  \begin{bmatrix}
    1+\frac{\Delta^2}{4} &(\pi_0-\pi_1)\frac{\Delta}{2}\\
    (\pi_0-\pi_1)\frac{\Delta}{2} & 1+2\pi_0\pi_1\Delta^2
  \end{bmatrix}
  \begin{bmatrix}
    1 \\
    \frac{\zeta}{\Delta}
  \end{bmatrix}\text{,}\\
Q_3&=\begin{pmatrix}
    1 & \frac{\zeta}{\Delta}
  \end{pmatrix}
  \frac{1}{A_0A_2-A_1^2}
  \begin{bmatrix}
    A_2 &A_1\\
    A_1 & A_0
  \end{bmatrix}
  \begin{bmatrix}
    1 \\
    \frac{\zeta}{\Delta}
  \end{bmatrix}. 
\end{split}
\end{equation*}
\end{theoremappendix}

\emph{Proof.} The proof can be found in \citet{efron1975efficiency}. \qed

\begin{theoremappendix}
If $\pi_i=\pi_j$, the expectation of the distance $d_{(i,j)}$ is a function of the Mahalanobis distance $\Delta_{i,j}$:

\begin{equation*}
\ep[d_{(i,j)}]=\sqrt{\frac{2}{\pi}}\exp\left( -\frac{\Delta_{i,j}^2}{8} \right) + \frac{1}{2}\Delta_{i,j}\left[ 1-2\Phi(-\frac{\Delta_{i,j}}{2}) \right]\text{,}
\end{equation*}

where $\Phi(\cdot)$ is the normal cumulative distribution function.
\end{theoremappendix}

\emph{Proof.} Since $d_{(i,j)}$ is the distance of $x_{(i)}$ to the decision boundary between class $i$ and $j$ decided by the Fisher's linear discriminant function $\lambda_{i,j}(x)=\beta_{i,j}+\alpha_{i,j}^\top x=0$, where

\begin{equation*}
\begin{split}
\beta_{i,j}&=\log(\pi_i/\pi_j)+\frac{1}{2}(\|\mu_j\|^2_2-\|\mu_i\|^2_2)\text{,}\\
\alpha_{i,j}^\top &=(\mu_i-\mu_j)^\top \text{.}
\end{split}
\end{equation*}

We have

\[d_{(i,j)}=\frac{\left|{\beta_{i,j}+\alpha_{i,j}^\top x_{(i)}}\right|}{\|\alpha_{i,j}\|_2}=\frac{\left|{\beta_{i,j}+\alpha_{i,j}^\top x_{(i)}}\right|}{\Delta_{i,j}}\text{.}\]

Because $x_{(i)}$ is sampled from the conditional Gaussian distribution of class $i$, there is

\[x_{(i)}\sim\mathcal{N}(\mu_i,I)\text{.}\]

Let $H_{i,j}=\beta_{i,j}+\alpha_{i,j}^\top x_{(i)}$, and $\zeta_{i,j}=\log(\pi_i/\pi_j)$, then according to the property of Gaussian distribution we can know that

\[H_{i,j}\sim\mathcal{N}(\mu'_{i,j},\sigma_{i,j}^2)\text{,}\]

where

\begin{equation*}
\begin{split}
\mu'_{i,j}&=\beta_{i,j}+\alpha_{i,j}^\top \mu_i\\
&=\zeta_{i,j}+\frac{1}{2}(\|\mu_j\|^2_2-\|\mu_i\|^2_2)+(\mu_i-\mu_j)^\top \mu_i\\
&=\zeta_{i,j}+\frac{1}{2}\|\mu_i-\mu_j\|_2^2\\
&=\zeta_{i,j}+\frac{1}{2}\Delta_{i,j}^2\text{,}\\
\end{split}
\end{equation*}

and $\sigma_{i,j}^2=\alpha_{i,j}^\top I \alpha_{i,j}=\Delta_{i,j}^2$. Thus $\left|H_{i,j}\right|$ distributes as a \emph{Folded Gaussian (Normal) Distribution}. From the property of folded Gaussian distribution we know that

\[\ep[\left|H_{i,j}\right|]=\sqrt{\frac{2}{\pi}}\sigma_{i,j}\exp(-\frac{{\mu'_{i,j}}^2}{2{\sigma_{i,j}}^2})+\mu'_{i,j}[1-2\Phi(-\frac{\mu'_{i,j}}{\sigma_{i,j}})]\text{.}\]

For notation clarity, we let

\[\alpha_{i,j}=\frac{\mu'_{i,j}}{\sigma_{i,j}}=\frac{1}{2}\Delta_{i,j}+\zeta_{i,j}/\Delta_{i,j}\text{.}\]

Since $\ep[d_{(i,j)}]=\ep[\left|H_{i,j}\right|]/\Delta_{i,j}$, we have

\begin{equation*}
\ep[d_{(i,j)}]=\sqrt{\frac{2}{\pi}}\exp(-\frac{\alpha_{i,j}^2}{2})+\alpha_{i,j}[1-2\Phi(-\alpha_{i,j})]\text{.}
\end{equation*}

The derivative of $\ep[d_{(i,j)}]$ to $\Delta_{i,j}$ is

\begin{equation*}
\begin{split}
\frac{\partial\ep[d_{(i,j)}]}{\partial\Delta_{i,j}}&=\frac{\partial\ep[d_{(i,j)}]}{\partial\alpha_{i,j}}\cdot\frac{\partial\alpha_{i,j}}{\partial\Delta_{i,j}}\\
&=[1-2\Phi(-\frac{1}{2}\Delta_{i,j}-\frac{\zeta_{i,j}}{\Delta_{i,j}})](\frac{1}{2}-\frac{\zeta_{i,j}}{\Delta_{i,j}^2})\\
&=[1-2\Phi(-\Delta_{i,j}(\frac{1}{2}-\frac{\zeta_{i,j}}{\Delta_{i,j}^2}))](\frac{1}{2}-\frac{\zeta_{i,j}}{\Delta_{i,j}^2})\\
&\geq0\text{,}
\end{split}
\end{equation*}

where the Mahalanobis distance $\Delta_{i,j}$ is non-negative. Specially, when $\pi_i=\pi_j$, i.e., $\zeta_{i,j}=0$, there is

\begin{equation*}
\ep[d_{(i,j)}]=\sqrt{\frac{2}{\pi}}\exp(-\frac{\Delta_{i,j}^2}{8})+\frac{1}{2}\Delta_{i,j}[1-2\Phi(-\frac{\Delta_{i,j}}{2})]\text{.}
\end{equation*}

\qed

\begin{theoremappendix}
Assume that $\sum_{i=1}^L\mu_i=0$ and $\left\|\mu \right\|_2^2=C$. Then there has
\begin{equation*}
\overline{\RB}\leq \sqrt{\frac{LC}{2(L-1)}}\text{.}
\end{equation*}
The equality holds if and only if
\begin{equation*}
\mu_i^\top \mu_j=\begin{cases}
C\text{,}& i=j\text{,}\\
C/(1-L)\text{,}& i\neq j\text{,}
\end{cases}
\end{equation*}
where $i,j\in[L]$ and $\mu_i,\mu_j \in \mu$.
\end{theoremappendix}

\emph{Proof.} According to the definition of $\overline{\RB}$, we have

\begin{equation*}
\begin{split}
\overline{\RB}&=\frac{1}{2}\min_{i,j\in[L]}\Delta_{i,j}\\
&=\frac{1}{2}\sqrt{\min_{i,j\in[L]}\Delta_{i,j}^2}\\
&\leq \frac{1}{2}\sqrt{\frac{1}{L(L-1)}\sum_{i\neq j}\Delta_{i,j}^2}\\
&=\frac{1}{2}\sqrt{\frac{1}{L(L-1)}\sum_{i\neq j}(\|\mu_i\|_2^2+\|\mu_j\|_2^2-2\mu^\top_i\mu_j)}\\
&=\frac{1}{2}\sqrt{\frac{2}{L}\sum_{i\in[L]}{\|\mu_i\|_2^2}-\frac{1}{L(L-1)}\sum_{i\neq j}2\mu^\top_i\mu_j}\\
&=\frac{1}{2}\sqrt{\frac{2}{L-1}\sum_{i\in[L]}{\|\mu_i\|_2^2}-\frac{1}{L(L-1)}(\sum_{i\in[L]}\mu_i)^2}\text{,}\\
\end{split}
\end{equation*}

Since $\sum_{i=1}^L\mu_i=0$ and $\left\|\mu \right\|_2^2=C$, we further have

\begin{equation*}
\begin{split}
\overline{\RB}&\leq \sqrt{\frac{1}{2(L-1)}\sum_{i\in[L]}{\|\mu_i\|_2^2}}\\
&\leq \sqrt{\frac{LC}{2(L-1)}}\text{.}
\end{split}
\end{equation*}

Note that the final equality holds if and only if all the equalities hold, i.e., there are

\begin{equation*}
\|\mu_i\|_2^2=C,\forall i\in[L]\text{,}
\end{equation*}

and

\begin{equation*}
\Delta_{i,j}=\text{constant},\forall i\neq j\text{.}
\end{equation*}

Thus we can easily derive that the final equality holds if and only if

\begin{equation*}
\mu_i^\top \mu_j=\begin{cases}
C\text{,}& i=j\text{,}\\
C/(1-L)\text{,}& i\neq j\text{,}
\end{cases}
\end{equation*}
where $i,j\in[L]$ and $\mu_i,\mu_j \in \mu$.

\qed

\section{More Discussions and Details}
In this section we discuss more on the loss function of the MM-LDA network, and the choice of the square norm of $C$ of MMD. Besides, we provide technical details of the adversarial training methods we use in our experiments.

\subsection{The Loss Function of the MM-LDA Network}
Considering that the network with parameters $\theta$ induces a joint distribution on the latent feature $z$ and the label $y$ as $Q_{\theta}(z,y)$. We denote the MMD as $P(z,y)$, $\HH(P,Q)$ as the cross-entropy for the distributions $P$ and $Q$. Then the training objective could be designed as

\begin{equation*}
\begin{split}
\HH(Q_{\theta},P)&=\ep_{(z,y)\sim Q_{\theta}}[-\log P(y|z)-\log P(z)]\\
&=\ep_{(z,y)\sim Q_{\theta}}[-\log P(y|z)]+\ep_{z\sim Q_{\theta}^{'}}[-\log P(z)]\text{.}
\end{split}
\end{equation*}

Here $Q_{\theta}^{'}$ is the marginal distribution of $Q_{\theta}$ for $z$. Since we are focusing on classification tasks, we assume for tractability that the marginal distribution $Q_{\theta}^{'}(z)$ is consistent with it of the MMD, i.e., $P(z)$. Therefore, minimizing $\HH(Q_{\theta},P)$ equals to minimizing $\ep_{(z,y)\sim Q_{\theta}}[-\log P(y|z)]$, which further leads to the loss function $\mathcal{L}_{\text{MM}}$ under the Monte Carlo approximation. In practice, the gap between $Q_{\theta}^{'}(z)$ and $P(z)$ would not influence the performance, as shown in our experiment results.

In order to better gather the latent feature vectors to their corresponding conditional Gaussian distributions, the loss function of the MM-LDA network should be

\[\mathcal{L}_{\text{MM}}=-F_{\text{MM}}(\mu_y^*)^\top\log F_{\text{MM}}(x)\text{,}\]

where $F_{\text{MM}}(\mu_y^*)_k=P(y=k|z=\mu_y^*),k\in[L]$. $F_{\text{MM}}(\mu_y^*)$ is the prediction vector on the mean vector $\mu_y^*$ in MMD. We further have

\begin{equation*}
\begin{split}
F_{\text{MM}}(\mu_y^*)_y&=\frac{\exp(2{\mu_y^*}^\top\mu_y^*)}{\sum_{i\in[L]}\exp(2{\mu_y^*}^\top\mu_i^*)}\\
&=\frac{1}{1+\sum_{i\neq y}\exp[2{\mu_y^*}^\top(\mu_i^*-\mu_y^*)]}\\
&=\frac{1}{1+(L-1)\exp(-\frac{2LC}{L-1})}\text{.}
\end{split}
\end{equation*}

Thus the $L_\infty$-distance between the one-hot label vector $1_y$ and $F_{\text{MM}}(\mu_y^*)$ is

\begin{equation*}
\begin{split}
\|1_y-F_{\text{MM}}(\mu_y^*)\|_\infty&\leq\left|1-F_{\text{MM}}(\mu_y^*)_y\right|\\
&=\frac{1}{1+\frac{1}{L-1}\exp(\frac{2LC}{L-1})}\text{.}
\end{split}
\end{equation*}

It is easy to see that the gap between $1_y$ and $F_{\text{MM}}(\mu_y^*)$ rapidly decreases w.r.t $C$. For instance, when $C>10$, $L=10$, we numerically have $\|1_y-F_{\text{MM}}(\mu_y^*)\|_\infty\leq 10^{-8}$. Therefore we use the one-hot label $1_y$ in the loss function because of its simplicity.

%

\subsection{Technical Details of Adversarial Training}
Our experiments are done on NVIDIA Tesla P100 GPUs. The number of the adversarial fine-tuning steps on MNIST is 10,000 and on CIFAR-10 is 30,000. We apply a constant learning rate of $0.01$ on both datasets. The mixing ratio of normal examples and adversarial examples is 1:1. Averagely the time cost to craft an adversarial example using FGSM, BIM and ILCM is less than $0.1$ seconds, while using JSMA is around $5$ seconds. This makes adversarial training on JSMA computationally expensive.

\end{document}